\documentclass[letterpaper, 10 pt, conference]{ieeeconf}  

\IEEEoverridecommandlockouts                              

\title{\LARGE \bf
Open3DTrack: Towards Open-Vocabulary 3D Multi-Object Tracking
}

\author{Ayesha Ishaq$^{1}$, Mohamed El Amine Boudjoghra$^{1}$, Jean Lahoud$^{1}$, \\ Fahad Shahbaz Khan$^{1,2}$, Salman Khan$^{1,3}$, Hisham Cholakkal$^{1}$, Rao Muhammad Anwer$^{1}$ 
\thanks{$^{1}$ Mohamed Bin Zayed University of Artificial Intelligence}
\thanks{$^{2}$ Linköping University, ~~~~~~ $^{3}$ Australian National University}
}

\usepackage{graphicx}
\usepackage{amsmath}
\usepackage{float}
\usepackage{xcolor} 
\definecolor{lightblue}{RGB}{55, 127, 185}
\definecolor{darkergreen}{RGB}{73, 135, 72}
\usepackage{colortbl}
 
\usepackage{pdfpages}
\usepackage{caption}
\usepackage{hyperref}
\begin{document}

\maketitle
\thispagestyle{empty}
\pagestyle{empty}

\begin{abstract}
3D multi-object tracking plays a critical role in autonomous driving by enabling the real-time monitoring and prediction of multiple objects' movements. Traditional 3D tracking systems are typically constrained by predefined object categories, limiting their adaptability to novel, unseen objects in dynamic environments. To address this limitation, we introduce open-vocabulary 3D tracking, which extends the scope of 3D tracking to include objects beyond predefined categories. We formulate the problem of open-vocabulary 3D tracking and introduce dataset splits designed to represent various open-vocabulary scenarios. We propose a novel approach that integrates open-vocabulary capabilities into a 3D tracking framework, allowing for generalization to unseen object classes. Our method effectively reduces the performance gap between tracking known and novel objects through strategic adaptation. 
Experimental results demonstrate the robustness and adaptability of our method in diverse outdoor driving scenarios. To the best of our knowledge, this work is the first to address open-vocabulary 3D tracking, presenting a significant advancement for autonomous systems in real-world settings. Code, trained models, and dataset splits are available at \href{https://github.com/ayesha-ishaq/Open3DTrack}{https://github.com/ayesha-ishaq/Open3DTrack}.

\end{abstract}

\section{INTRODUCTION}

3D multi-object tracking involves detecting and continuously tracking multiple objects in the physical space across consecutive frames. 
This task is essential in autonomous driving as it allows the vehicle to monitor and predict the movements of multiple objects in its environment. The system can make informed decisions for safe navigation, collision avoidance, and trajectory planning by accurately identifying, localizing, and tracking objects over time. This capability is vital for real-time responsiveness and the overall reliability of autonomous systems.

Current 3D tracking systems are structured around benchmarks with labeled datasets, driving the development of methods that excel at tracking predefined categories such as cars, pedestrians, and cyclists \cite{10341778, 3dmotformer, 9578166,liu2022gnn,chiu2021probabilistic,benbarka2021score}. These tracking systems maximize performance within this closed set by detecting and associating known objects across frames. However, these systems are limited by their reliance on a closed set of classes, making them less adaptable to new or unexpected objects in dynamic real-world environments. Although effective within defined benchmarks, their performance diminishes when encountering unknown objects.

To overcome these limitations, open-vocabulary systems are needed to enable the detection and recognition of objects beyond the predefined categories. In 3D tracking, open-vocabulary capabilities are essential for enhancing adaptability and robustness in dynamic environments, where a vehicle may encounter unexpected or rare objects not covered by the training data. By incorporating open-vocabulary approaches, 3D tracking systems can better generalize to new objects, ensuring safer and more reliable autonomous navigation in diverse real-world scenarios.

\begin{figure}
    \centering
    \begin{tabular}{c}
    Known class track: \textcolor{red}{Pedestrian} \\
    \hspace{-0.5cm}
    \includegraphics[trim={0 114cm 0 13cm},clip,width=1\linewidth]{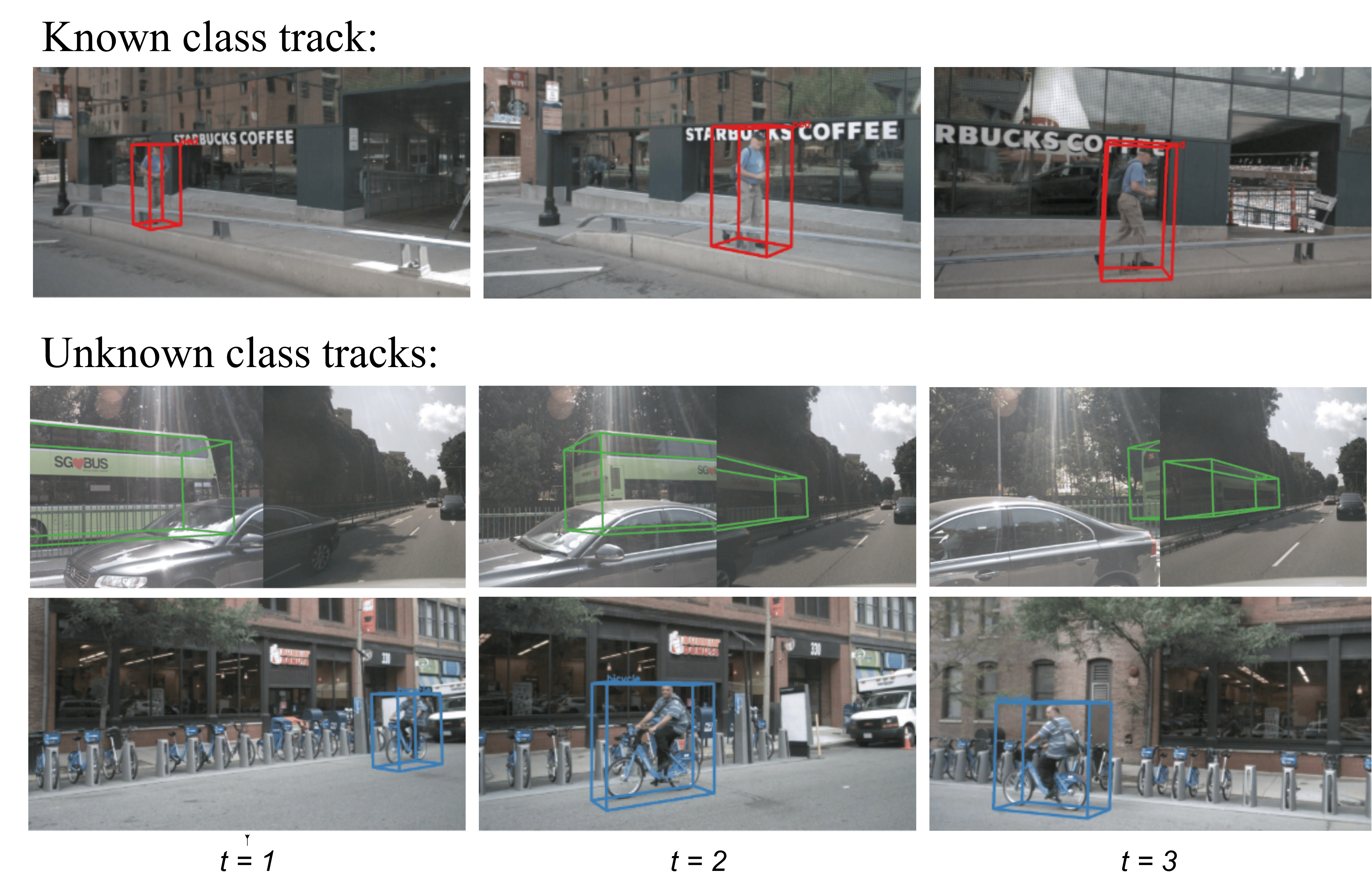}\\
    Unknown class track: \textcolor{darkergreen}{Bus}, \textcolor{lightblue}{Bicycle} \\
    \hspace{-0.5cm}
    \includegraphics[trim={0 1cm 0 75cm},clip,width=1\linewidth]{figures/novel.drawio.png}
    \end{tabular}
    \caption{\textbf{Open3DTrack} We train the tracking system on classes from $\mathcal{C}^{base}$ and prompt it with unseen categories $\textit{bicycle, bus} \in \mathcal{C}^{novel}$ at test time. Here, we show the tracked 3D bounding boxes projected on multiple frames and views. Our method can track and label both known and unknown object classes across consecutive 3D frames.}
    \label{fig:bicycle_novel}
\end{figure}

In this paper, we introduce the task of open-vocabulary 3D tracking, where trajectories of known and unknown object classes are estimated in the real 3D space by linking their positions across consecutive frames.
We formulate the open-vocabulary tracking problem, propose splits to evaluate the performance across known and unknown classes, and introduce a method that bridges the performance gap between tracking known and unseen classes. 
Our approach leverages
2D open-vocabulary methods and 3D tracking frameworks to generalize to unseen categories effectively.
Through extensive experiments, we demonstrate that our method achieves strong results through our adaptation strategies, providing a robust solution for open-vocabulary 3D tracking. Fig. \ref{fig:bicycle_novel} shows an example output of our solution. 
Our method can detect and track previously unseen object categories in three-dimensional space without relying on predefined labels.

Our contributions can be summarised as follows:
\begin{itemize}
    \item We introduce the task of open-vocabulary 3D tracking that generalizes to unseen categories.
    \item We propose a novel problem formulation and evaluation splits for open-vocabulary tracking.
    \item We develop a method that effectively bridges the performance gap between known and unknown class tracking.
    \item Our extensive experimental validation shows the high effectiveness of our proposed adaptation.
\end{itemize}

\section{Related Work}
Our work focuses on 3D tracking and open-vocabulary systems. This section reviews existing approaches in these areas, highlighting key techniques and advancements to provide an understanding of current methodologies.

\subsection{3D Tracking}
In 3D object tracking, various approaches have been explored, leveraging different features and methodologies. Traditional approaches rely on Kalman filters or other probabilistic models, with recent advances shifting towards deep learning-based methods. The tracking-by-detection paradigm remains dominant, particularly in 3D multi-object tracking (MOT). Model-based techniques, such as AB3DMOT \cite{9341164}, extend traditional Kalman filters to 3D for motion estimation, coupling it with the Hungarian method to provide a standard 3D MOT baseline. Another method, EagerMOT \cite{9562072}, combines observations from LiDAR and camera sensors, enhancing the ability to track distant objects and achieve precise trajectory localization. The Immortal Tracker \cite{wang_chen_pang_wang_zhang_2022} introduces a trajectory prediction mechanism using a Kalman filter to maintain tracks even when objects temporarily disappear from view, significantly reducing identity switches. CenterPoint \cite{9578166} and CenterTrack \cite{9523000} combine velocity estimates with point-based detection to track the centers of objects across multiple frames. Poly-MOT \cite{10341778} introduces a method that adapts a tracking criterion based on object categories, employing category-specific motion models and a two-stage data association strategy. In 3DMOTFormer \cite{3dmotformer}, the authors introduce a geometry-based 3D multi-object tracking framework built on the transformer architecture, using an Edge-Augmented Graph Transformer to perform data association through edge classification.
While great advancement has been made in the 3D tracking task, all these methods rely on tracking a known predefined set of object classes.

\subsection{Open-Vocabulary 3D}
Open vocabulary 3D tasks are emerging in computer vision, driven by the need to generalize beyond predefined categories, particularly in dynamic environments. Recent advances in 2D open vocabulary tasks \cite{ViLD, Lseg, openseg, owlv2} have inspired the extension of these concepts to the 3D domain, where the complexity of object shapes, occlusions, and sparse point-cloud data pose unique challenges. Introducing open-vocabulary techniques in 3D vision enables flexible, scalable models that identify diverse objects without extensive labeling.
PointCLIP \cite{9878980} leverages the Contrastive Vision-Language Pre-training (CLIP) model to achieve open-vocabulary point cloud recognition by aligning CLIP-encoded point clouds with 3D category texts through knowledge transfer from 2D images to 3D. For 3D open-vocabulary tasks, using image features obtained from vision language models and aligning them with point cloud features is common \cite{10204547}, \cite{10377184}. OpenScene \cite{10203983} generates dense features for 3D points embedded within the CLIP feature space by using image-per-pixel features from 2D semantic segmentation models \cite{openseg}. OpenMask3D \cite{openmask3d} introduces a zero-shot approach for open-vocabulary 3D instance segmentation by utilizing class-agnostic 3D instance masks and aggregating per-mask features through multi-view fusion of CLIP-based image embeddings.
Beyond semantic segmentation, in \cite{OV3DDet}, the authors propose combining a point cloud detector for object localization with cross-modal contrastive learning to connect an image, point cloud, and text representations, enabling open-vocabulary 3D detection in indoor scenes. Open-YOLO 3D \cite{openyolo3d} has also explored combining class-agnostic 3D proposals with strong 2D open-vocabulary detectors, demonstrating improvements in both accuracy and speed.
However, open-vocabulary 3D tracking remains largely unexplored, presenting a significant gap in leveraging these emerging techniques for dynamic, real-time object tracking in the physical space.

\begin{figure*}[ht]
    \centering
    \includegraphics[width=1\linewidth]{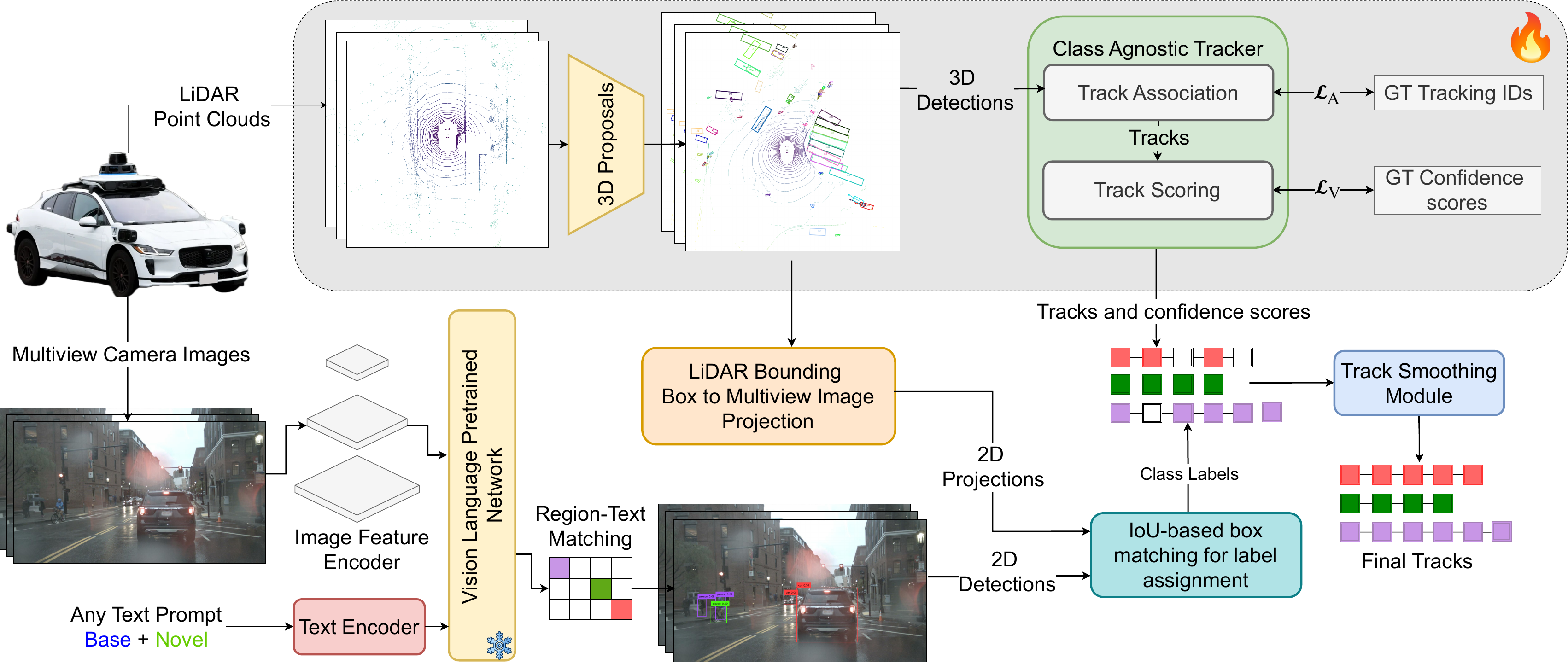}
    \caption{\textbf{System overview.} Open3DTrack leverages 3D proposals from base classes $C^{base}$ to train the 3D tracker in a class-agnostic manner, enabling it to construct tracks and predict confidence scores. During inference, the system classifies 3D proposals using open-vocabulary categories from both base and novel classes $ C^{base} \cup C^{novel} $, utilizing 2D image cues and pretrained vision-language model. This process labels the tracks output by the tracker, effectively generating open-vocabulary 3D tracks.}
    \label{fig:ov3dtrack}
\end{figure*}

\section{Methodology}
In this section, we first formulate the problem of 3D multi-object tracking in an open-vocabulary setting. We then describe our approach for adapting an existing tracker to a class-agnostic open-vocabulary tracker. Finally, we outline the dataset splits used for evaluation.
\subsection{Problem Formulation}
Given a sequence of LiDAR frames $\mathbf{P} = \{P_t\}_{t=1}^T$, where $P_t$ represents the point cloud at time $t$, the goal of 3D multi-object tracking is to detect and track multiple objects $\{O_i^t\}_{i=1}^{N_t}$, where $O_i^t$ denotes the $i$-th object and $N_t$ are the total objects in frame $t$.
Each $O_i^t$ comprises the object class $c$ and the 3D bounding box $b=[x,y,z,w,l,h,\theta]$ where $x,y,z$ represent the coordinates of the center of the bounding box, and $w,l,h$ are the width, length, and height of the bounding box, and $\theta$ is the heading angle, respectively.

In an open-vocabulary setting, we train a tracker $\mathcal{T}$ on $\mathbf{P}^{train}$ and their corresponding annotation $A^{train}$ of objects with semantic categories $\mathcal{C}^{base}$. At test time, we aim to find tracks $G_T$ of objects in $\mathbf{P}^{test}$ belonging to $\mathcal{C}^{base}$ as well as a given $\mathcal{C}^{novel}$, where $\mathcal{C}^{base} \cap \mathcal{C}^{novel} = \phi$. 
In a tracking-by-detection framework, only class labels and scores for $\mathcal{C}^{base}$ object detections are available at train and test time. 
Our setup requires the tracker $\mathcal{T}$ to track arbitrary object classes with $\mathcal{C}^{novel}$ not appearing during training time.

\subsection{Open-Vocabulary 3D Multi-Object Tracking}

In this section, we present our approach for 3D open-vocabulary tracking. As illustrated in Fig. \ref{fig:ov3dtrack}, our proposed system integrates 3D proposals, 2D image cues, and vision-language models to classify and track objects in an open-vocabulary setting, including classes not encountered during training.

\subsubsection*{\textbf{3D Tracker}}
We adapt our 3D open-vocabulary tracking approach from the recent 3D multi-object tracker 3DMOTFormer. \cite{3dmotformer}.
3DMOTFormer is a tracking-by-detection framework that leverages graph structures to represent relationships between existing tracks and new detections. Features such as position, size, velocity, class label, and confidence score are processed through a graph transformer, and updated based on interactions. Within the decoder, edge-augmented cross-attention models how tracks and detections interact, with edges representing potential matches. 
To match tracks with detections, affinity scores are predicted which indicate the likelihood of correspondence. The final step computes loss using positive target edges that share the same tracking ID. Ground truth tracking IDs are assigned by calculating 3D IoU between ground truth boxes and 3D detections, followed by Hungarian Matching for one-to-one ID assignment. Loss \(\mathcal{L}_m\) accumulated over the sequence is used to train the network with back-propagation through time (BPTT). The tracker is trained on a closed set of classes, producing tracks $G_T$ of objects from those classes.  

While 3DMOTFormer performs well for the closed set, it struggles to track unseen object classes due to its reliance on class-specific information during training. The tracker expects identified 3D boxes to generate edge proposals and then process those through the graph transformer. In a situation where it may receive a new label or an unlabelled object, the model will fail to track it. 3DMOTFormer also uses class-specific statistics within its pipeline, such as maximal class velocities, to compute association edges between tracks and new detections. Such information regarding unseen objects, that the system may encounter for the first time, is not available in a real-world open-vocabulary setting.

\subsubsection*{\textbf{Class-Agnostic Tracking}}
 
To enable tracking irrespective of object class label, we modify the tracker to utilize only object position, dimensions, heading angle, and velocity as the initial features. Object detections from a given detector are treated as proposals (class-agnostic) where the class labels and confidence scores are dropped and not utilized.

To compute potential association edges, we replace class-specific distance thresholds with a fixed threshold, achieving a balance between maintaining true edges and avoiding unnecessary associations. We change the GT track ID assignment by computing 3D IoU between ground truth boxes $A^{train}$ and 3D detections belonging to $\mathcal{C}^{base}$ in a class-agnostic manner.

 \begin{figure*}[ht!]
    \centering
    \includegraphics[width=1\linewidth]{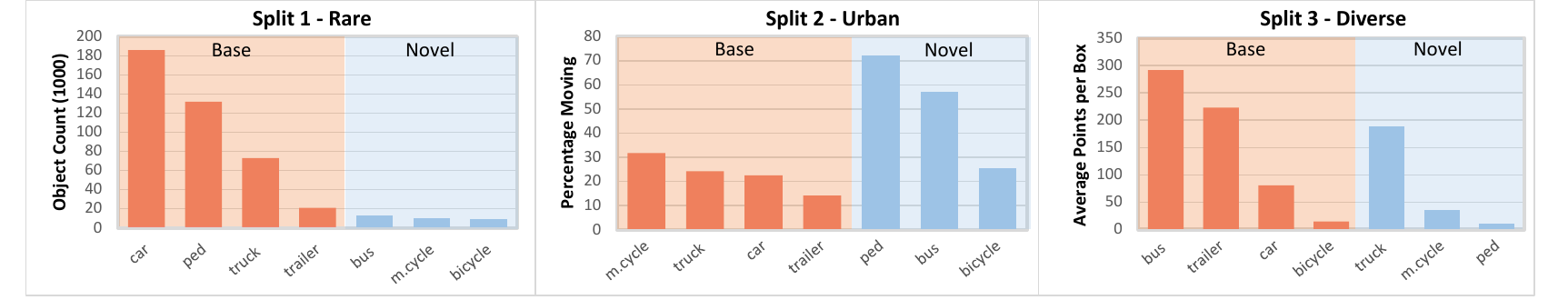}
    \caption{\textbf{Open-vocabulary dataset splits}. Our proposed dataset splits for known and unknown categories are based on statistics from the nuScenes\cite{nuscenes2019} dataset, which are shown here. The first split considers the least occurring objects as novel. In the second split, we analyze the motion patterns of objects in urban settings, motivating our urban vs highway split. Novel classes in this split exhibited the highest percentage of movement. In the final split, we use the average points per box to ensure a similarly diverse object distribution in the novel split in terms of object volume.}
    \label{fig:splits}
\end{figure*}

\subsubsection*{\textbf{Confidence Score Prediction}}
\label{CSP}
As discussed earlier, we drop class labels and confidence scores from the detection features to ensure the tracker is independent of any information specific to object classes. 
As a result, the initial output from the system suffers from a lack of representation of the objectness score of each box. To overcome this issue, we use the detector objectness scores for $\mathcal{C}^{base}$ and train a regression head at the end of the tracker. The regression head is composed of a feed-forward network that takes as input processed detection features $h^{(Ld)}_{D,j}$ from the graph transformer and predicts a single confidence score. We employ a straightforward Mean Squared Error loss between predicted and true confidence scores to learn the scores for the detection boxes. The combined loss is 
$\mathcal{L} = \mathcal{L}_m + \lambda_c \mathcal{L}_c$, where $\mathcal{L}_c$ is the loss for confidence score, weighted by $\lambda_c$.

\subsubsection*{\textbf{2D Driven Open Vocabulary Labels}}
While we train the tracker by removing all class-specific dependencies, we require a strong open-vocabulary classification for all object tracks, known and unknown, at test time. 
We leverage vision-language models to match regions in the images with text prompts, enabling it to detect and classify objects for which it has not been explicitly trained. 
2D detections on multi-view images from the vehicle are obtained from an open-vocabulary 2D detector, which is prompted with all possible labels—both $\mathcal{C}^{base}$ and $\mathcal{C}^{novel}$. To enhance the cross-modal alignment between visual and textual information, we employ a YOLO-based framework, where a CLIP Encoder transforms input text into text embeddings while a darknet backbone extracts multi-scale image features. These representations are then refined using the RepVL-PAN module, which combines image and text embeddings to enable region-text matching.

The system allocates class labels to the 3D proposals by first projecting the 3D boxes onto multi-view images. The labels are then derived from the highest IoU overlap with 2D detections on the image plane. Since multi-view images provide a 360-degree view, a 3D box may project onto multiple images. A single projection is used, selecting its largest rectangular image region for IoU matching to obtain its 2D equivalent. Projections of the 3D detections that do not overlap with a 2D open-vocabulary detection are labeled 'unknown'. We explicitly identify these boxes at the end of the tracking process. 

\subsubsection*{\textbf{Track Consistency Scoring}}
At the output of the tracker, we apply a track smoothing module that addresses the issue of unknown boxes that do not match any 2D detections from the images, as well as improve label consistency throughout tracks to account for inaccuracies in the open-vocabulary detector outputs and errors accrued during projection from 3D to 2D. 
To calculate the most accurate class for each track, we first compute the weight of an object based on its detected bounding box in an image, with objects at a greater distance assigned a smaller weight. The depth is derived from the size of the bounding box, the object's vertical position in the image for perspective correction, and the aspect ratio of the bounding box to ensure uniformity for all classes. Size is the normalized box area, given by:
    \(
   B_{size} = \frac{B_{width} \times B_{height}}{I_{width} \times I_{height}}
   \)
   \[
    depth = 
   \frac{1}{B_{size}}(1 - \alpha_{p} \left( \frac{y_{center}}{I_{height}} \right))
   \]

   \[
  depth = 
   \begin{cases} 
  depth \times \left( \frac{1}{aspect\_ratio} \right) & \text{if } aspect\_ratio > \beta_{ar} \\
   depth  & \text{otherwise}
   \end{cases}
   \]
   
   \noindent where,
   $B$ refers to 2D box, $I$ refers to image, $y_{center}$ points to the vertical center of the box, $\alpha_p$ is the perspective correction factor estimated by comparing true depth to calculated depth, while $\beta_{ar}$ is the aspect ratio threshold selected by analyzing the distribution across common objects in outdoor scenes, and \(aspect\_ratio = \frac{B_{width}}{B_{height}}\).

We then compute the weight as \(w_{dist}= e^{\frac{-depth}{\lambda_s}}\) where $\lambda_s$ is a depth scaling factor. The predicted confidence score of each box is multiplied with $w_{dist}$, yielding higher confidence for boxes that appear closer in images. 
The smoothing module then uses the modified confidence score and finds the average modified confidence for each class in the track, along with their occurrence, selecting the class with the highest combination. Any track that is composed entirely of unknowns is dropped.

\subsection{Open-Vocabulary Tracking Splits}
\label{data_splits}

Our method is evaluated in outdoor scenes using custom data splits based on common object types and situations. Fig. \ref{fig:splits} shows various statistics from nuScenes \cite{nuscenes2019} and subsequent splits. One split follows the common open-vocabulary approach, treating rarer classes as novel. Another split differentiates between highway and urban objects based on the greater activity 
of some classes within urban areas. A final split focuses on objects with high variability
, represented by the average LiDAR points per object. These splits provide a robust framework for evaluating open-vocabulary tracking in realistic and diverse outdoor settings.

\section{Experiments}
\begin{table*}[ht!]
\centering
\caption{Results on the nuScenes \cite{nuscenes2019} validation set using 3D proposals from CenterPoint \cite{9578166}. We compare the baseline and our open-vocabulary method across various splits. Baseline without unknowns drops any 3D proposals unmatched to 2D detections. Closed-set 3DMOTFormer results are shown for reference. Performance is reported for primary metrics AMOTA$\uparrow$ and AMOTP$\downarrow$. Highlighted cells represent novel classes.}
\resizebox{\textwidth}{!}{%
\begin{tabular}{lllllllllllllllll}
\hline
 &
  \multicolumn{2}{c}{\textbf{Overall}} &
  \multicolumn{2}{c}{\textbf{Bicycle}} &
  \multicolumn{2}{c}{\textbf{Bus}} &
  \multicolumn{2}{c}{\textbf{Car}} &
  \multicolumn{2}{c}{\textbf{Motorcycle}} &
  \multicolumn{2}{c}{\textbf{Pedestrian}} &
  \multicolumn{2}{c}{\textbf{Trailer}} &
  \multicolumn{2}{c}{\textbf{Truck}} \\ \hline
 &
  AMOTA &
  AMOTP &
  \multicolumn{1}{c}{AMOTA} &
  \multicolumn{1}{c}{AMOTP} &
  \multicolumn{1}{c}{AMOTA} &
  \multicolumn{1}{c}{AMOTP} &
  \multicolumn{1}{c}{AMOTA} &
  \multicolumn{1}{c}{AMOTP} &
  \multicolumn{1}{c}{AMOTA} &
  \multicolumn{1}{c}{AMOTP} &
  \multicolumn{1}{c}{AMOTA} &
  \multicolumn{1}{c}{AMOTP} &
  \multicolumn{1}{c}{AMOTA} &
  \multicolumn{1}{c}{AMOTP} &
  \multicolumn{1}{c}{AMOTA} &
  \multicolumn{1}{c}{AMOTP} \\ \hline
\multicolumn{17}{c}{\textbf{Split 1 - Rare}} \\ \hline
\begin{tabular}[c]{@{}l@{}}Baseline   \\ (w/o unknowns)\end{tabular} &
 0.415 &
  0.868 &
  \cellcolor[HTML]{EFEFEF}0.323 &
  \cellcolor[HTML]{EFEFEF}1.051 &
  \cellcolor[HTML]{EFEFEF}0.403 &
  \cellcolor[HTML]{EFEFEF}1.030 &
  0.656 &
  0.432 &
  \cellcolor[HTML]{EFEFEF}0.194 &  
  \cellcolor[HTML]{EFEFEF}1.336 &
  0.717 &
  0.399 &
  0.395 & 
  1.069 &
  0.218 & 
  0.758 \\
\begin{tabular}[c]{@{}l@{}}Baseline\\ (w/ unknowns)\end{tabular} &
  0.483 &
  0.785 &
  \cellcolor[HTML]{EFEFEF}0.424 &
  \cellcolor[HTML]{EFEFEF}0.853 &
  \cellcolor[HTML]{EFEFEF}0.423 &
  \cellcolor[HTML]{EFEFEF}0.952 &
  0.678 &
  0.391 &
  \cellcolor[HTML]{EFEFEF}0.288 &
  \cellcolor[HTML]{EFEFEF}1.293 &
  0.746 &
  0.350 &
  0.469 &
  0.975 &
  0.349 &
  0.684 \\
\begin{tabular}[c]{@{}l@{}}\textbf{Open3DTrack} \\ \textbf{(Ours)}\end{tabular} &
  0.578 &
  0.783 &
  \cellcolor[HTML]{EFEFEF}0.445 &
  \cellcolor[HTML]{EFEFEF}0.988 &
  \cellcolor[HTML]{EFEFEF}0.612 &  
  \cellcolor[HTML]{EFEFEF}0.918 &
  0.779 & 
  0.390 &
  \cellcolor[HTML]{EFEFEF}0.469 &
  \cellcolor[HTML]{EFEFEF}1.104 &
  0.752 &  
  0.403 &
  0.477 &  
  1.037 &
  0.511 &  
  0.638 \\ \hline
\multicolumn{17}{c}{\textbf{Split 2 - Urban}} \\ \hline
\begin{tabular}[c]{@{}l@{}}Baseline\\ (w/ unknowns)\end{tabular} &
  0.469 &
  0.740 &
  \cellcolor[HTML]{EFEFEF}0.301 &
  \cellcolor[HTML]{EFEFEF}0.982 &
  \cellcolor[HTML]{EFEFEF}0.481 &
  \cellcolor[HTML]{EFEFEF}0.907 &
  0.657 &
  0.392 &
  0.537 &
  0.549 &
  \cellcolor[HTML]{EFEFEF}0.482 &
  \cellcolor[HTML]{EFEFEF}0.634 &
  0.455 &
  0.985 &
  0.367 &
  0.729 \\
\begin{tabular}[c]{@{}l@{}}\textbf{Open3DTrack}\\ \textbf{(Ours)}\end{tabular} &
  0.590 &
  0.677 &
  \cellcolor[HTML]{EFEFEF}0.400 &
  \cellcolor[HTML]{EFEFEF}0.894 &
  \cellcolor[HTML]{EFEFEF}0.683 &
  \cellcolor[HTML]{EFEFEF}0.820 &
  0.788 &
  0.387 &
  0.702 &
  0.422 &
  \cellcolor[HTML]{EFEFEF}0.548 &
  \cellcolor[HTML]{EFEFEF}0.599 &
  0.488 &
  0.981 &
  0.522 &
  0.635 \\ \hline
\multicolumn{17}{c}{\textbf{Split 3 - Diverse}} \\ \hline
\begin{tabular}[c]{@{}l@{}}Baseline \\ (w/ unknowns)\end{tabular} &
  0.438 &
  0.813 &
  0.389 &
  0.493 &
  0.640 &
  0.499 &
  0.627 &
  0.402 &
  \cellcolor[HTML]{EFEFEF}0.275 &
  \cellcolor[HTML]{EFEFEF}1.311 &
  \cellcolor[HTML]{EFEFEF}0.564 &
  \cellcolor[HTML]{EFEFEF}0.590 &
  0.424 &
  0.993 &
  \cellcolor[HTML]{EFEFEF}0.144 &
  \cellcolor[HTML]{EFEFEF}1.406 \\
\begin{tabular}[c]{@{}l@{}}\textbf{Open3DTrack}\\ \textbf{(Ours)}\end{tabular} &
  0.536 &
  0.804 &
  0.524 &
  0.581 &
  0.770 &
  0.543 &
  0.708 &
  0.395 &
  \cellcolor[HTML]{EFEFEF}0.438 &
  \cellcolor[HTML]{EFEFEF}1.143 &
  \cellcolor[HTML]{EFEFEF}0.564 &
  \cellcolor[HTML]{EFEFEF}0.648 &
  0.470 &
  1.036 &
  \cellcolor[HTML]{EFEFEF}0.276 &
  \cellcolor[HTML]{EFEFEF}1.281 \\ \hline
\multicolumn{17}{l}{} \\ \hline
\begin{tabular}[c]{@{}l@{}}Upper Bound \\ (3DMOTFormer)\end{tabular} &
  0.710 &
  0.521 &
  0.545 &
  0.429 &
  0.853 &
  0.527 &
  0.838 &
  0.382 &
  0.723 &
  0.459 &
  0.812 &
  0.335 &
  0.509 &
  0.936 &
  0.690 &
  0.576 \\ \hline
\end{tabular}%
}
\label{tab:table_main}
\end{table*}

\begin{table}[ht]
\centering
\caption{Results on the nuScenes \cite{nuscenes2019} validation set for detector generalization for open-vocabulary 3D tracking. Results are shown for base and novel class groups from split 3 \ref{data_splits}, with the tracker trained on detections from the source method and tested on 3D detections from the target method. }
\resizebox{0.45\textwidth}{!}{%
\begin{tabular}{llccccc}
\hline
            &             & \multicolumn{2}{c}{Base} & \multicolumn{2}{c}{Novel} & \multicolumn{1}{c}{Overall}\\ \hline
Source      & Target      & AMOTA       & AMOTP      & AMOTA       & AMOTP  & AMOTA     \\ \hline
Centerpoint & Centerpoint & 0.630       & 0.617    & 0.509     & 1.003    &   0.578    \\
MEGVII      & MEGVII      & 0.586	  &  0.7055     & 0.477       & 1.1207 & 0.539    \\
BEVFusion   & BEVFusion   & 0.662	  & 0.5535     & 0.410       & 1.054  & 0.554     \\ \hline
Centerpoint & MEGVII      & 0.579       & 0.7106    & 0.496        & 1.093  &  0.543   \\
Centerpoint & BEVFusion   & 0.616      & 0.556     & 0.458       & 0.982 &  0.548\\ \hline
\end{tabular}%
}
\label{tab:detectors}
\end{table}
\vspace{-1em}
\subsection{Datasets}
We use the nuScenes \cite{nuscenes2019} dataset for training and evaluation of our open-vocabulary tracking system. NuScenes features data collected from several sensors, such as 360-degree cameras, a 32-beam LiDAR, and RADARs annotated at 2Hz. It includes 1000 scenes, each lasting 20 seconds, and is split into 700 scenes for training, 150 for validation, and 150 for testing. It offers seven unique classes for the tracking task.

\subsubsection*{Evaluation Metrics}

We evaluate our results using the nuScenes \cite{nuscenes2019} tracking benchmark with AMOTA and AMOTP as the primary metrics \cite{amota}. AMOTA averages MOTA across recall levels, accounting for false positives, missed targets, and identity switches, with lower impact at recall levels below 10\%. AMOTP measures the average alignment error between predicted and annotated bounding boxes, reflecting localization accuracy across all frames.

\subsection{Implementation Details}

We follow 3DMOTFormer’s default settings, training for 12 epochs with AdamW, a batch size of 8, and a learning rate of 0.001 with 0.01 weight decay. The regression loss weight $\lambda_c$ is set to 0.5. In the association graph, the distance threshold for edge truncation is 3 meters. For open-vocabulary 2D detection, we use the YOLOv8L-Worldv2 model \cite{Cheng2024YOLOWorld} with a confidence threshold of 0.01. In the Track Consistency Scoring module, perspective correction $\alpha_p$ is 0.2, aspect ratio threshold $\beta_{ar}$ is 2.5, and depth scale $\lambda_s$ is 250.

\begin{table}[ht]
\centering
\caption{Ablation study on the effect of each component of our proposed method.}
\resizebox{0.45\textwidth}{!}{%
\begin{tabular}{lcccc}
\hline
Component                                  & \multicolumn{2}{c}{Base} & \multicolumn{2}{c}{Novel} \\ \hline
                                           & AMOTA     & AMOTP     & AMOTA     & AMOTP     \\ \hline
Baseline                                   & 0.504     & 0.6638    & 0.482     & 0.841     \\
+ Class-Agnostic GT assignment             & 0.536     & 0.6215    & 0.437     & 0.804     \\
+ Confidence Score Prediction              & 0.618     & 0.617     & 0.507     & 0.7987    \\
+ Track Consistency Scoring                & 0.625     & 0.6063    & 0.544     & 0.771     \\ \hline
\end{tabular}%
}
\label{tab:ablation}
\end{table}

\subsection{Benchmark Results}
\label{results}

\begin{figure*}[ht!]
    \includegraphics[trim={0 2cm 0 2cm},clip,width=1\linewidth]{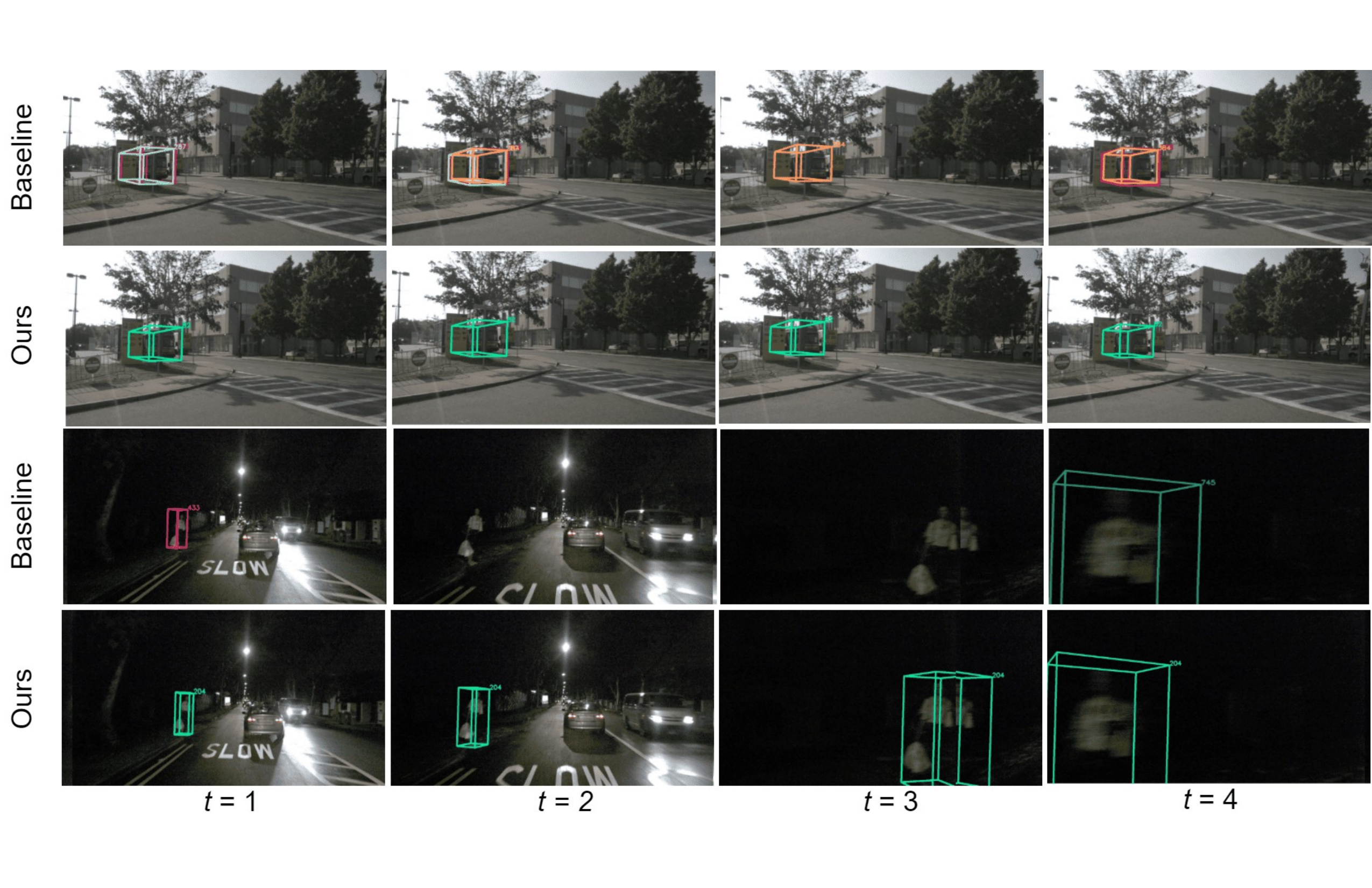}
    \caption{\textbf{Qualitative comparison.} We show the output of our method and the baseline on challenging examples from nuScenes. The top two rows show an occluded novel class object, \textit{bus}, which is tracked successfully by ours, while it changes IDs in the baseline output. The bottom two rows show an example of poor lighting and blurred novel object, \textit{pedestrian}, tracked by ours but generating a fragmented track by the baseline. Additional results are presented in the accompanying video.}
    \label{fig:qualitative}
\end{figure*}

Table \ref{tab:table_main} shows results on the nuScenes \cite{nuscenes2019} validation set using 3D proposals from CentrePoint \cite{9578166} detector. We compare our open-vocabulary method on the various proposed splits with the baseline. For reference, we show the original closed-set results of 3DMOTFormer as an upper bound. The baseline method is a simple adaptation of 3DMOTFormer, where we replace class labels with labels obtained from 2D open-vocabulary detections and make use of confidence scores from the 2D detections as well. 
For the first split, we show results for baseline with and without unmatched detections. With our presented adaptations, we achieve an overall AMOTA of 0.567 in the first split, with strong performance scores for the novel classes, improving significantly on top of the baseline. In each of the last two splits, we achieve a similar improvement over the baseline, with an overall AMOTA of 0.59 in split 2 and 0.536 in split 3.
Particularly in split 3, we observe AMOTP increasing across some classes. This may be due to the Track Consistency Scoring, where unknown 3D proposals might be incorrectly identified, increasing location error but improving AMOTA due to better track continuity. 
We show some challenging novel class examples in Fig. \ref{fig:qualitative}.
\\

We evaluate our method's generalization across different detectors,  CenterPoint\cite{9578166}, MEGVII \cite{megvii} and BEVFusion \cite{bevfusion}, as shown in Table \ref{tab:detectors}. Our approach delivers consistent performance across various 3D proposal sources.
The best results are from the standard detector, likely due to alignment between train and test bounding boxes. Interestingly, BEVFusion gives the highest base class performance and the lowest novel class performance, possibly due to training on high-quality base class proposals and less noise for regulation.

\section{Ablation}
In this section, we detail the effects of various components of our proposed system. All experiments reported are on the nuScenes validation set using CenterPoint detections and specifically on split 2 from section \ref{data_splits}. Table \ref{tab:ablation} displays the improvements with each component.

\textbf{Baseline: }As detailed in section \ref{results}, baseline for our system is simply 3DMOTFormer trained on labels obtained from 2D open-vocabulary detections. All other class-specific details were removed. A simple track smoothing method is applied to the results.

\textbf{Class-Agnostic GT Assignment: }To build on top of our baseline we assign GT tracking IDs to 3D proposals using 3D IoU, omitting class labels. This improves results for the base classes at test time, while results drop slightly for novel classes. This occurs because class-agnostic GT assignment improves target accuracy as compared to noisy open-vocabulary base class labels, leading to better learning on known classes while reducing the regularization and dropping performance on unknown classes. However, the overall improvement in performance is significant.

\textbf{Confidence Score Prediction: }As discussed in section \ref{CSP}, the confidence scores from the 2D detections did not capture the true objectness score of the 3D proposals resulting in lower performance. There was also no estimation of confidence scores for unmatched proposals. To overcome these issue we added the confidence prediction head. We see a significant boost from this module in both the base and novel classes, 8.2\%P and 7\%P respectively

\textbf{Track Consistency Scoring: }Simple track smoothing is necessary to assign labels to unknown detections, however, we found that formalizing a mechanism to score the identified detections and use these scores to attain consistency in track labels considerably improved performance for all categories. The overall AMOTA increased by 2\%P upon the addition of this module.

 \section{Conclusion}
 In conclusion, we introduce a novel approach for open-vocabulary 3D multi-object tracking, addressing closed-set limitations. By leveraging 2D open-vocabulary detections and applying strategic adaptations to a 3D tracking framework, such as class-agnostic tracking and confidence score prediction, we improve tracking novel object classes. Our experiments highlight the robustness of our method. These contributions represent a meaningful step toward more flexible 3D tracking solutions for autonomous systems, enhancing their ability to generalize in dynamic environments.

 \textbf{Acknowledgment}
 The computations were enabled by resources provided by NAISS at Alvis partially funded by Swedish Research Council through grant agreement no. 2022-06725, LUMI hosted by CSC (Finland) and LUMI consortium, and by Berzelius resource provided by the Knut and Alice Wallenberg Foundation at the NSC.


\bibliographystyle{IEEEtran}
\bibliography{main}
\end{document}